\documentclass[conference]{IEEEtran}
\IEEEoverridecommandlockouts
\usepackage{cite}
\usepackage{amsmath,amssymb,amsfonts}
\usepackage{algorithmic}
\usepackage{graphicx}
\usepackage{textcomp}
\usepackage{xcolor}
\usepackage{marvosym}
\usepackage{balance}
\def\BibTeX{{\rm B\kern-.05em{\sc i\kern-.025em b}\kern-.08em
    T\kern-.1667em\lower.7ex\hbox{E}\kern-.125emX}}

\usepackage{fancyhdr}

\begin{document}

\fancyhf{} 
\fancyhead[C]{This article has been accepted by ICME 2023 (IEEE International Conference on Multimedia and Expo 2023).}

\title{Low-Light Image Enhancement by Learning Contrastive Representations in Spatial and Frequency Domains\\

\thanks{{*} represents co-first authorship.

         {\Letter} represents the corresponding author.}
}

\author{\IEEEauthorblockN{1\textsuperscript{st} Yi Huang}
\IEEEauthorblockA{\textit{School of Information Engineering,} \\
\textit{Southwest University of Science and Technology}\\
Mianyang, China \\
yihuangswust@outlook.com}\\

 \IEEEauthorblockN{2\textsuperscript{nd} Gui Fu}
\IEEEauthorblockA{\textit{School of Avionics and Electrical Engineering} \\
\textit{Civil Aviation Flight College of China}\\
Guanghan, China \\
abyfugui@163.com}\\
\and
\IEEEauthorblockN{1\textsuperscript{st} Xiaoguang TU}
\IEEEauthorblockA{\textit{School of Avionics and Electrical Engineering} \\
\textit{Civil Aviation Flight College of China}\\
Guanghan, China \\
xguangtu@outlook.com}\\

\IEEEauthorblockN{3\textsuperscript{rd} Tingting Liu}
\IEEEauthorblockA{\textit{School of Information Engineering,} \\
\textit{Southwest University of Science and Technology}\\
Mianyang, China \\
liutingting@swust.edu.cn}
\and
\IEEEauthorblockN{4\textsuperscript{th} Bokai Liu}
\IEEEauthorblockA{\textit{School of Avionics and Electrical Engineering} \\
\textit{Civil Aviation Flight College of China}\\
Guanghan, China \\
bk liu1991@163.com}
\and
\IEEEauthorblockN{5\textsuperscript{th} Ming Yang}
\IEEEauthorblockA{\textit{School of Avionics and Electrical Engineering} \\
\textit{Civil Aviation Flight College of China}\\
Guanghan, China \\
yangming932@163.com}

\and
\IEEEauthorblockN{6\textsuperscript{th} Ziliang Feng*}
\IEEEauthorblockA{\textit{School of Computer Science} \\
\textit{Sichuan University}\\
Chengdu, China \\
fengziliang@scu.edu.cn}
}

\maketitle

\begin{abstract}
Images taken under low-light conditions tend to suffer from poor visibility, which can decrease image quality and even reduce the performance of the downstream tasks. It is hard for a CNN-based method to learn generalized features that can recover normal images from the ones under various unknow low-light conditions. In this paper, we propose to incorporate the contrastive learning into an illumination correction network to learn abstract representations to distinguish various low-light conditions in the representation space, with the purpose of enhancing the generalizability of the network. Considering that light conditions can change the frequency components of the images, the representations are learned and compared in both spatial and frequency domains to make full advantage of the contrastive learning. The proposed method is evaluated on 
 LOL and LOL-V2 datasets, the results show that the proposed method achieves better qualitative and quantitative results compared with other state-of-the-arts.
\end{abstract}


\section{Introduction}
Images collected under low-light conditions have problems such as low brightness, unsaturated colors, and missing details, which can lower image quality and hence limit the performance of downstream tasks. In the past decades, a number of methods have been proposed to recover a high-quality image from the one under low-light condition. These methods can be generally divided into two categories, i.e., the Classical and Learning-based methods, respectively.
\begin{figure}
	\centering 
	\includegraphics[height=5cm,width=8.5cm]{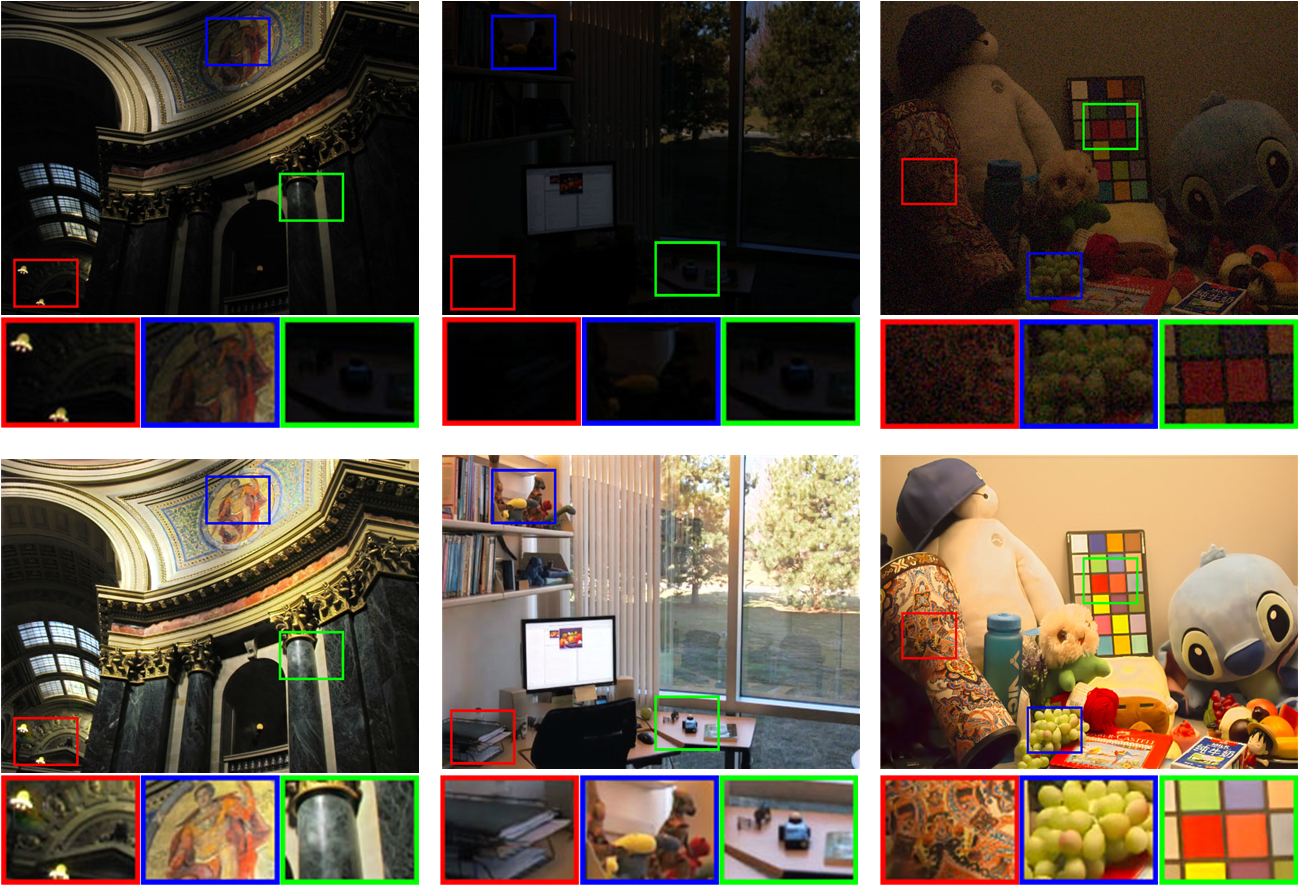}
	\caption{Illumination correction results of our model. Images in the first row are the low-light inputs, images in the second row are the correction results.}
       \label{1}
\end{figure}
\begin{figure*}
	\centering 
	\includegraphics[height=5.3cm,width=17.5cm]{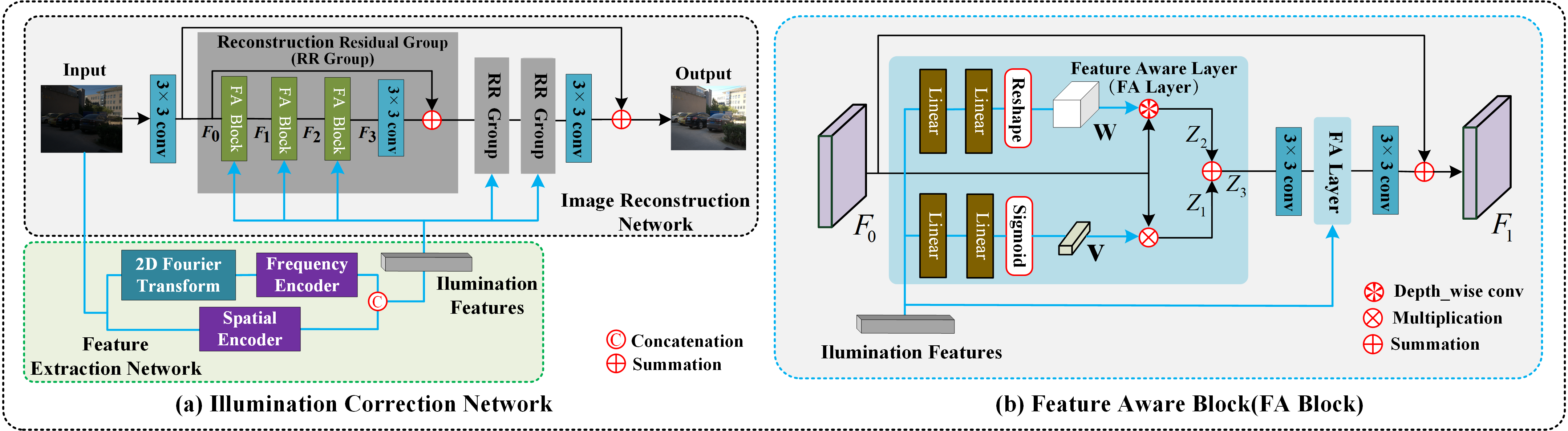}
	\caption{The structure of our method. Figure (a) is the overall illumination correction network, which is composed of feature extraction network and image reconstruction network. Figre (b) is the FA block in the image reconstruction network.}
              \label{2}
\end{figure*}

Classical methods generally use Histogram Equalization(HE)~\cite{kim1997contrast} or Retinex-based methods~\cite{wang2013naturalness} to improve the quality of the low-light images. HE improves the global brightness of the images by making the histogram conform to the uniform distribution. However, it ignores the local brightness difference of the images, easy to result in local over-enhancement and noise amplification. Retinex-based methods decompose the image into reflection component and illumination component, and then remove the illumination intensity from the illumination component separately. Methods in this category always appear with a pre-defined regularization term, making the model hard to be optimized, where a large number of hyperparameters need to be manually adjusted.

Recently, with the increasing computational power of computers~\cite{liu2022novel,liu2022distributed,liu2022design}, the CNN-based illumination correction methods have achieved promising performance in terms of both image quality and algorithm efficiency. Shen et al.~\cite{shen2017msr} proposed the earliest CNN-based model for image illumination correction. It combines Retinex theory and CNN to design a multi-scale encoder MSR-net, which enhances the brightness of the image by decomposing it in advance. Lore et al.~\cite{lore2017llnet} designed a low-light autoencoder, which for the first time taking into account noise removal and detail recovery. Different from Retinex-based methods, Zhu et al.~\cite{zhu2020eemefn} proposed a new type of three-branch convolutional network, dividing the image into three parts: illumination, reflection and noise, refining the image components to accurately correct the image. Such CNN-based algorithms have improved the performance to a great extent, but it is difficult for them to learn generalized features directly from supervised learning, since the publicly available datasets are incomplete for training. 

To address this issue, we propose a novel CNN-based illumination correction framework by taking advantage of the constrative learning. The proposed network is designed to distinguish features of images under various low-light conditions in the representation space, rather than directly supervised by the labeled training data. During such a constrative training process, the network would pay more attention to feature differences instead of creating a straightforward mapping from the input to the output images. Hence, the generalizability of the network can be improved. In addition, the previous works~\cite{mildenhall2021nerf} have discovered that the CNN-based methods tend to reconstruct low-frequency components in image reconstruction tasks, where part of the high-frequency components will be dropped, resulting in the inevitable spectral bias problem. To alleviate this issue, we used a weight-adaptive loss function to increase the network's attention to high-frequency signals. The results of the proposed method is shown in Fig.\ref{1}, where all images are recovered to very clear ones even the input images are almost black. 

Our main contributions are sumarized as follows:
\begin{itemize}
	\item An novel network based on contrastive learning is proposed to recover normal images from the ones under various low-light conditions, The network learns how to distinguish different low-light features represented in the spatial and frequency domains by conducting contrastive learning.
	\item During the image correction process a weight-adaptive frequency loss is utilized to constrain the network, which increases the network's focus on the reconstruction of high-frequency signals in the image, thereby addressing the issue of spectral bias.
	\item The proposed method shows a pleasing performance in the task of illumination correction, the low-light images are recovered to very high-quality ones, the generaliability of the network are significantly improved. 
\end{itemize}

\section{Related work }
In this section, we give a brief review for the contrastive learning and frequency domain learning.

Contrastive learning~\cite{wu2018unsupervised} is one of the most competitive feature learning methods for unlabeled samples. It trains the encoder by controlling the distance of different samples in the feature space. Specifically, in feature space, query sample features attract similar positive sample features and reject dissimilar negative sample features. The well-designed training target InfoNCE~\cite{oord2018representation} can represent the mutual information of two similar samples. When InfoNCE training reaches convergence, it is considered that the query sample features are basically the same as the positive sample features. 

Reasonable selection of positive (similar) and negative (dissimilar) pairs of samples is one of the important design aspects of contrastive learning models. Positive samples can be multiple perspectives of the same thing~\cite{wang2021unsupervised}, transformers~\cite{wu2018unsupervised}, patches~\cite{oord2018representation}, videos with different time steps~\cite{zhuang2020unsupervised}. Negative samples can be randomly selected images. Contrastive learning methods have achieved satisfactory results in different tasks. Wang et al.~\cite{wang2021exploring} proposed cross-image pixel contrastive learning in image segmentation tasks. Li et al.~\cite{li2020prototypical} combined clustering and contrastive learning and applied it to transfer tasks. Commonly, contrastive learning is also heavily used in tasks such as machine translation~\cite{yang2019reducing}, dialogue~\cite{cai2020group}, etc.

In the task of image reconstruction, a gap between the generated image spectrum and the standard image spectrum can be observed. Such a phenomenon is referred to as spectral bias~\cite{mildenhall2021nerf}, showing that the CNN networks tend to concentrate on the low frequency components during image reconstruction or generation. Xu et al.~\cite{xu2019frequency} discover that the fitting priorities for different frequencies are different during network training, usually decreasing from low to high frequencies. Therefore, it is difficult for a CNN model to fully simulate all the frequency components for a standard image, especially those high frequency components that represents the edge information of the image.To overcome the spectral bias, Durall et al.~\cite{durall2020watch} adds a new spectral regularization term to train the CNN network. Tancik et al.~\cite{tancik2020fourier} map Fourier features into a low-dimensional feature domain to learn high-frequency features. Cho et al.~\cite{cho2021rethinking} minimize the $L_1$ distance between generated and standard images in the frequency domain. Although extensively methods have been explored, the spectral bias issue has not yet been overcome.

Inspired by previous works, we propose to incorporate the contrastive leaning into the illumination network to enhance the generalizability of the model. Moreover, we employ a weight-adaptive frequency loss to shrink feature distance from opposite classes in the frequency domain, aiming to address the issue of spectral bias during image reconstruction. 
\section{ Method}
\subsection{Model Structure}
 As shown in Fig.\ref{2}~(a), the overall illumination correction network consists of a \textbf{F}eature \textbf{E}xtraction \textbf{N}etwork (FEN) and an \textbf{I}mage \textbf{R}econstruction \textbf{N}etwork (IRN). The former extracts illumination features of an image in spatial and frequency domains, the later is made up of the reconstruction residual groups to reconstruct the low-light images to the normal ones.

\textbf{Feature extraction network}:
The FEN consists of a 2D Fourier transform layer and a two-stream encoder. The encoders perform pretraining through constrastive learning by a process as shown in Fig.\ref{3}. The query sample (orange box) is a randomly cropped patch in one of the input images. Other patches from the same images as the query sample are taken as the positive sample (yellow box), other patches from the different images are taken as the negative samples (green box). 
{Then, the patches are fed into our two-stream encoder, one stream for feature extracting in spatial domain, the other for feature extracting in frequency domain. Each encoder consists of 6 convolutional layers with an average pooling layer and an MLP layer. For frequency learning, patches are firstly transformed by a 2D Fourier transform. After that, patches in both domains are fed into the two separate encoders. Finally, the encoded spatial and frequency features are concatenated. The query, positive, and negative samples are encoded as $q$, $k_+$ and $k_-$, respectively. Following the MoCo~\cite{he2020momentum}, $q$ and $k_+$ are pulled close (similar), while $q$ and $k_-$ are pushed away (dissimilar) in feature space by the InfoNCE loss function:
\begin{equation}
	L_{Info} = -log \frac{exp(q\cdot k_+/\tau)}{\sum_{j=1}^{queue} exp(q\cdot k_j/\tau)}
\end{equation}

where `queue' represents the number of negative samples, and $\tau$ is a hyperparameter that represents the punishment on the hard negative samples. If $\tau$ is larger, the punishment is weaker. The symbol $\cdot$ represents the dot product of vectors. 

\begin{figure}
	\centering 
	\includegraphics[height=3.3cm,width=8.5cm]{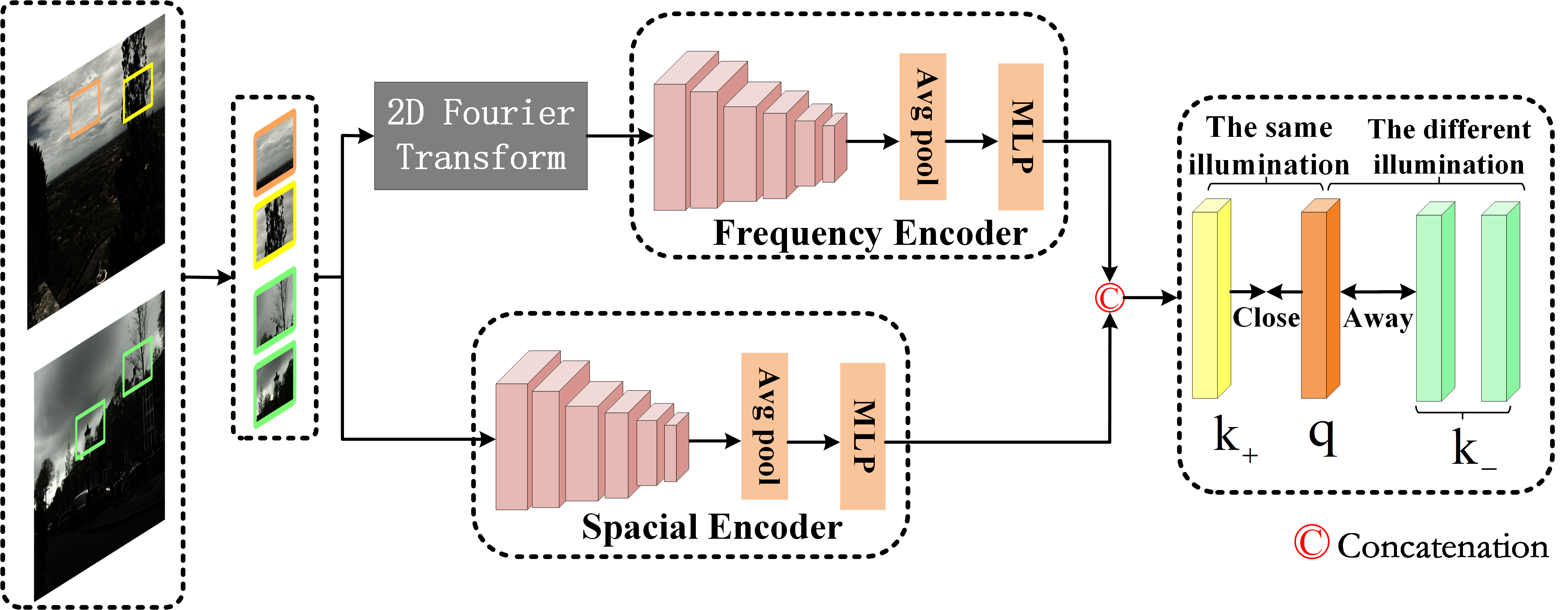}
    
	\caption{Training process for our two-stream encoder. The encoders are trained in a contrastive learning manner.}
         \label{3}
\end{figure}
\textbf{Image reconstruction network}:
The IRN (Fig.\ref{2}~(a)) are made up by \textbf{F}eature \textbf{A}ware block (FA block) and $3\times3$ convolution layers with residual structure. The whole construction network contains three residual groups, each of which contains three FA blocks and a $3\times3$ convolution, respectively.

The FA block is shown in Fig.\ref{2}~(b). Each FA block consists of two \textbf{F}eature \textbf{A}ware layers (FA layer), and two $3\times3$ convolutions, respectively. The illumination features are used to generate the reconstruction kernel $w\in R^{3 * 3 * C}$  and channel modulation coefficient $v$ through the FA layer. 
The shallow feature ($F_0$) of the image is convolved with $w$ to generate $Z_2$. In addition, the weight of different channels of $F_0$ is scaled by $v$ to obtain $Z_1$. $Z_1$ is summed up with $Z_2$ and fed into the subsequent layers to produce the output feature $F_1$. The $F_2$ and $F_3$ of Fig.\ref{2}~(a) can be obtained by repeating the appeal operation. After all operations of Fig.\ref{2}~(a) are completed, the normal light image can be reconstructed.
\subsection{Loss Function}
The overall loss functions are shown in formula (2), where $L_1$ represents the reconstruction loss, and $L_{fre}$ represents the frequency loss function, respectivley. $L_{Info}$ is the constrative learning loss, which is used to fine-tune FEN. $w_1$ and $w_2$ are the weight parameters. 
\begin{equation}
	Loss = L_{Info} + w_1\cdot L_1 + w_2\cdot L_{fre}
\end{equation}

The loss function in spatial domain adopts $L_1$ norm:
\begin{equation}
	L_1(\hat{y}_i,y_i) = \frac{1}{n} \sum_{i=1}^n |y_i-\hat{y}_i|
\end{equation}

Where $\hat{y}_i$ and $y_i$ are the pixel values of reconstructed images and the standard images, respectively. 

In order to address the spectral bias of convolutional neural network, we employ a frequency loss whose weight can be automatically adjusted. By increasing the weight of high-frequency loss, the network can pay more attention to high-frequency components. The pixel value $f(x,y)$ in the spatial domain is converted into the frequency domain value $F(u,v)$ according to the 2D Fourier transform (Formula 4):
\begin{equation}
	F(u,v) = \sum_{x=0}^{M-1} \sum_{y=0}^{N-1} f(x,y)\cdot e^{-i2\pi(\frac{ux}{M}+\frac{vy}{N})}
\end{equation}

Let $F_r(u,v)$  and  $F_f(u,v)$ be single frequency components corresponding to the standard image and the reconstructed image, respectively. Then, calculating all the single component variances of the image to obtain the frequency loss function of formula(6), which adopts the focal loss idea of~\cite{jiang2021focal} and adds weight $w(u,v)$ before the frequency loss function. The specific solution of  $w(u,v)$ is shown in formula (5),  where $\alpha$ is the scaling factor. Since the reconstruction error is larger for the components of higher frequency, the weight $w(u,v)$ for the high-frequency components would be large, hence the model will pay more attention to the high-frequency components. In this way, the spectral bias issue can be alleviated.

\begin{align}
	& w(u,v) = |F_r(u,v)-F_f(u,v)|^\alpha \\
	& L_{fre} = \frac{1}{MN}\sum_{u=0}^{M-1}\sum_{v=0}^{N-1} w(u,v)|F_r(u,v)-F_f(u,v)|^2
\end{align}
\begin{figure}
	\centering 
	\includegraphics[height=5cm,width=8.5cm]{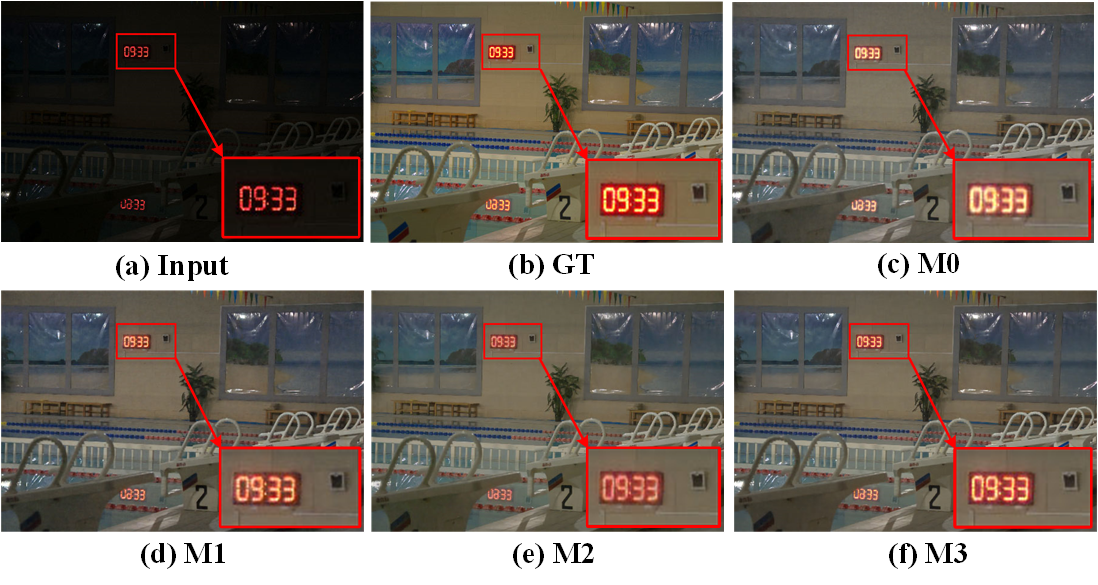}
 
	\caption{Component analysis results. (a) is the input image, (b) is the ground truth, (c)-(f) are the results of other settings.}
       \label{4}
\end{figure}
\begin{figure*}
	\centering 
	\includegraphics[height=5.1cm,width=17.5cm]{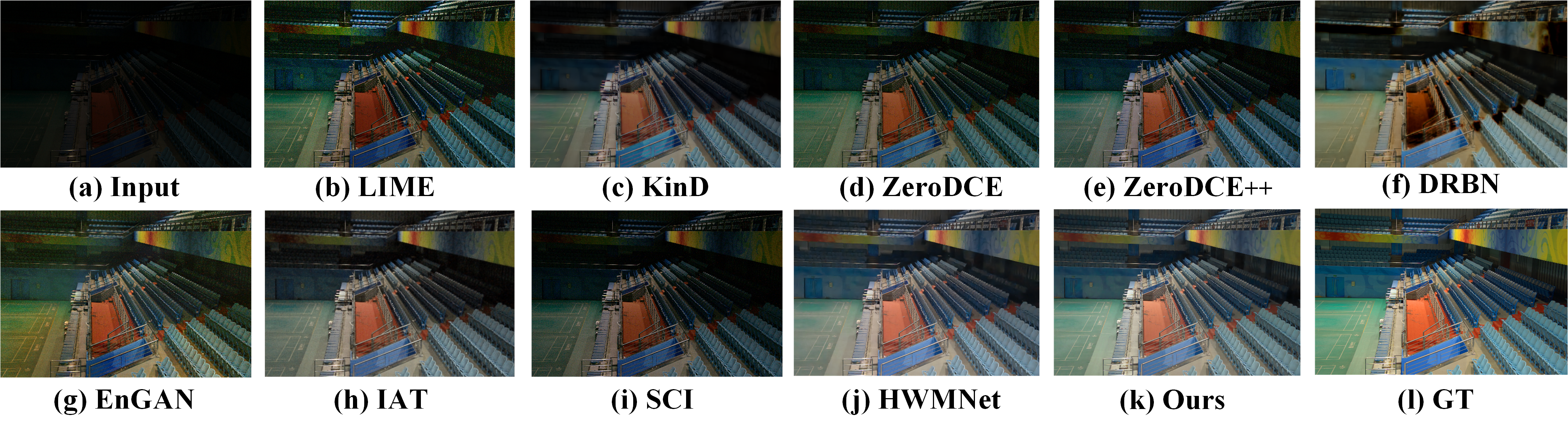}
	\caption{Visual comparison results for inter-dataset test. (a) is the input image, (k) is the result of our method, (l) is the ground truth, (b)-(j) are the results for other methods.}
      \label{5}
\end{figure*}
\section{expriment}
\subsection{Experiment Setup }
\begin{table}[t]
	\begin{center}
		\caption{Quantitative results for component analysis} \label{tab:cap}
  \setlength{\tabcolsep}{4mm}{
		\begin{tabular}{c c c c}
			\hline
			Combination name & PSNR $\uparrow$ & SSIM $\uparrow$ & LPIPS $\downarrow$
			\\
			\hline
			M0 & 22.52 & 0.83 & 0.15 \\
                M1 & 23.60 & 0.82 & 0.14 \\
			M2 & 21.82 & \textbf{0.84} & \textbf{0.12} \\
		      M3 &\textbf{24.25} & \textbf{0.84} & \textbf{0.12} \\
			\hline
		\end{tabular}
  }
	\end{center}
\end{table}
\textbf{Implementation Detail.} We use the LOL~\cite{wei2018deep} dataset to evaluate the proposed method with a NVIDIA GTX 3090Ti GPU. The PIL package is used for image augmentation during contrastive learning, where gamma correction and logarithmic adjustment are used to change image brightness, with randomly changed standard deviations for the Gaussian filters. The image size is $400\times600$, the patch size is $192\times192$. The scaling factor of the frequency loss function $\alpha=0.01$. The two-stream encoder is pre-trained for 200 times, and then the illumination correction network is trained for 500 times to complete all training. The Adam optimizer is adopted, the batch size is set to 16, the initial learning rate is set to 0.001 with a warm-up setting.

\textbf{Evaluation Metrics.} Three objective indexes, i.e., Peak Signal to Noise Ratio (PSNR), Structural Similarity (SSIM) and Learned Perceptual Image Patch Similarity (LPIPS) are used to evaluate the restored images. A higher PSNR value indicates a better performance, better SSIM is usually close to 1, and lower LPIPS value indicate better human perception.
\subsection{Ablation Study}
 We evaluate each of the proposed components for our method, the contrastive learning and freuquency domain learning, respectively. To this end, we evaluate four settings for the loss function combinations., i.e., $M0=L_1$, $M1=L_1+L_{fre}$,  $M2=L_1+L_{Info}$ and $M3=L_1+L_{fre}+L_{Info}$ respectively. $M0$ and $M1$ settings do not contain contrastive loss functions, and the two-stream encoder is not pre-trained in a contrastive learning manner. Other training parameters remain unchanged. The results of different combination settings are shown in Fig.\ref{4} and Tab.1.

\begin{table}[t]
	\begin{center}
		\caption{Quantitative results for inter-data test.} \label{tab:cap}
  \setlength{\tabcolsep}{5mm}{
		\begin{tabular}{c c c c}
			\hline
			Model & PSNR $\uparrow$ & SSIM $\uparrow$ & LPIPS $\downarrow$
			\\
			\hline
                ZeroDCE~\cite{guo2020zero} & 14.56 & 0.54 & 0.33 \\  
			LIME~\cite{guo2016lime}   & 16.77 & 0.56 & 0.35 \\   
			EnGAN~\cite{jiang2021enlightengan}  & 17.48 & 0.65 & 0.32 \\  
                RUAS~\cite{liu2021retinex}   & 18.23 & 0.72 & 0.35 \\
			KinD~\cite{zhang2019kindling}   & 20.87 & 0.80 & 0.17 \\
			DRBN~\cite{yang2020fidelity}   & 20.13 & 0.83 & 0.16 \\  
			KinD++~\cite{zhang2021beyond} & 21.30  & 0.82 & 0.16 \\
			IAT~\cite{Cui_2022_BMVC}    & 23.38 & 0.81 & 0.13 \\  
			MIRNet~\cite{zamir2020learning} & 24.14 & 0.83 & 0.13 \\
			Ours         &\textbf{24.25} & \textbf{0.84} & \textbf{0.12} \\
			\hline
		\end{tabular}
  }
	\end{center}
\end{table}

Fig.\ref{4} shows that the corrected images of $M0$ $\&$ $M1$ have obvious artifacts, the results of $M1$ $\&$ $M2$ have dark brightness. However, the image generated by $M3$ is very close to the standard image. Tab.1 shows that when both $L_{fre}$ and $L_{Info}$ are absence ($M0$), the performance is the worst. If the contrastive loss is missing ($M1$), the SSIM value is bad, indicating the illumination are not well removed. If the frequency loss is absence ($M2$), the noise are not well suppressed, causing the lower PSNR values. The ablation results shows that both the two-stream contrastive learning and the frequency loss function are effective for illumination correction.

\subsection{Comparison with state-of-the-arts}
\subsubsection{Inter-test results}
 We first evaluate our method on the LOL real dataset, where the default protocol splits the dataset into a training set and a testing set, respectively. We use the training set to train our method, and use the testing set for evaluation. Fig.\ref{5} shows the comparison results of our method with other state-of-the-art approaches on a very dark input image. It is clear to observe that our method achieves the best visualization result, which is closest to the ground truth. For the methods LIME~\cite{guo2016lime}, DRBN~\cite{yang2020fidelity}, ZeroDCE~\cite{guo2020zero}, KinD~\cite{zhang2019kindling}, it seems they can not well restore the input to a normal one. Moreover, the corrected results of EnGAN~\cite{jiang2021enlightengan}, ZeroDCE++~\cite{li2021learning}, IAT~\cite{Cui_2022_BMVC} contains obvious noise. Even HWMNet~\cite{fan2022half} achieves a similar good visualization result with ours, our method performs better on texture restoration. 

The quantitative results are shown in Tab.2, where we can seen our method achieves the best PSNR and SSIM values, indicating that our method performs better on noise removal and structure recovering. Our method also achieves the lowest LPIPS value, meaning the quality of images recovered by our method are better than those of other methods. 

\subsubsection{Cross-test results}
\begin{table}[t]
	\begin{center}
		\caption{Quantitative results for cross-dataset test.} \label{tab:cap}
        \setlength{\tabcolsep}{5mm}{
		\begin{tabular}{c c c c}
			\hline
			Medel & PSNR $\uparrow$ & SSIM $\uparrow$ & LPIPS $\downarrow$
			\\
			\hline
            RUAS~\cite{liu2021retinex}   & 15.35 & 0.50 & 0.31 \\
            ZeroDCE~\cite{guo2020zero}  & 18.10 & 0.58 & 0.32 \\
            DRBN~\cite{yang2020fidelity}   & 18.42 & 0.76 & 0.26 \\
            EnGAN~\cite{jiang2021enlightengan}   & 18.68 & 0.68 & 0.31 \\
		LLFlow~\cite{wang2022low} & 24.15 & 0.89 & 0.09 \\
		HWMNet~\cite{fan2022half}    & 30.30 & 0.90 & \textbf{0.08} \\
		Ours         & \textbf{30.89} & \textbf{0.91} & 0.09 \\
			\hline
		\end{tabular}
        }
	\end{center}
\end{table}
We perform a cross-dataset test to further evaluate the generalizability of our method, where LOL dataset is used for training and LOL-V2~\cite{yang2021sparse} is used for testing. One challenging images are selected to compare our method with others in Fig.\ref{6}. As can be seen, the results of EnGAN~\cite{jiang2021enlightengan} and ZeroDCE~\cite{guo2020zero} still contains a lot of noise and local overexposure. For LLFlow~\cite{wang2022low}, the results have serious color deviation, and it is not realistic enough. Our method achieves the results almost the same with the ground truth, demonstrating the superiority of our method in generating high-quality, visually pleasant images compared with other methods on the task of illumination correction.  Quantitative comparison results are listed in Tab.3, where we can see our method and HWMNet~\cite{fan2022half} achieve the best results on PSNR and SSIM metris, ranking the first and second, respectively. For the LPIPS metric, HWMNet~\cite{fan2022half} performs slightly better than our method (0.08 vs. 0.09).  

Through the cross-dataset testing results in Fig.6 and Tab.3, it is easy to conclude that our method achieves stronger generalizability than other methods such as  RUAS~\cite{liu2021retinex},  ZeroDCE~\cite{guo2020zero}, DRBN~\cite{yang2020fidelity},  EnGAN~\cite{jiang2021enlightengan}, and LLFlow~\cite{wang2022low}. Our method performs even slight better than the newly published method HWMNet~\cite{fan2022half}.   
\begin{figure}
	\centering 
	\includegraphics[height=5cm,width=8.5cm]{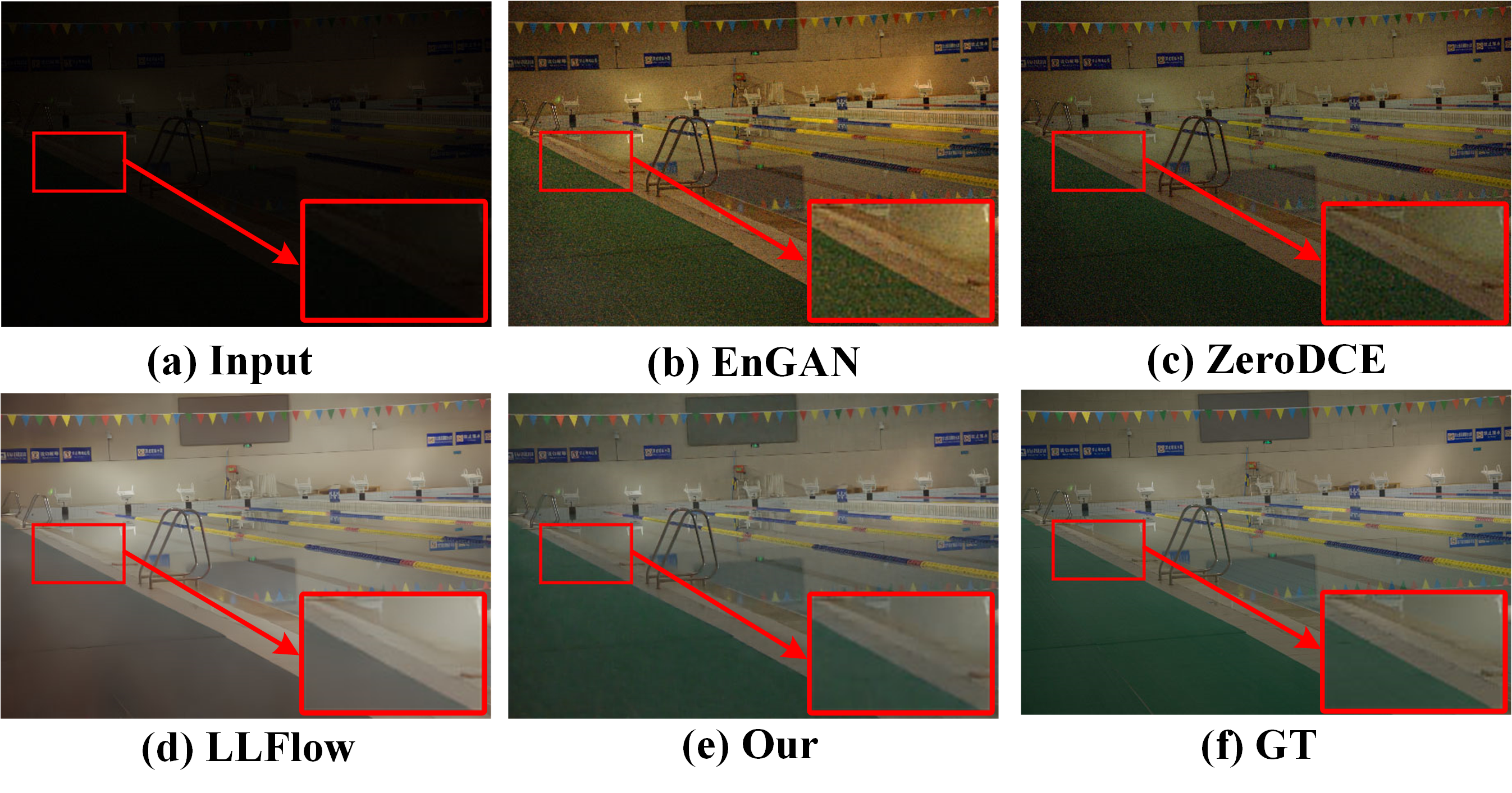}
	\caption{Visual comparison results for cross-dataset test. (a) is the input image, (e) is the result of our method, (f) is the ground
truth, (b)-(d) are the results for other methods.}
        \label{6}
\end{figure}
\section{Conclusion}
In this paper, an illumination correction model based on constrative learning is presented. Instead of designing a straitforward supervised training model, we propose to learn the differences of image features in representation space. A two-stream encoder with an adaptive weight frequency loss is designed to encode feature in both spatial and frequency domains. The experiment results show that our method can well restore the images to normal ones, even the inputs are visually very dark.  The visual results have justified the superiority of our method in generating high-quality and visually pleasant images, the cross-dataset testing results show our method has stronger generalizability than other methods. 
\section{Acknowledgement}
This work was supported in part by the Science and Technology Department in Sichuan Province of China under Grant 2022JDRC0076, in part by China postdoctoral science foundation under Grant 2022M722248, in part by the Project of Basic Scientific Research of Central Universities of China under Grant ZHMH2022-004 and J2022-025, and in part by Open Fund of Key Laboratory of Flight Techniques and Flight Safety, CAAC (No. FZ2022KF06). 

\balance
\bibliographystyle{IEEEtran}
\bibliography{conf}
\end{document}